\documentclass{sig-alternate}

\usepackage{graphicx}
\usepackage{caption}
\usepackage{subcaption}
\usepackage{multirow}
\usepackage{url}
\usepackage{balance}
\def\etal{\emph{et al.}~}
\DeclareMathOperator*{\argmax}{arg\,max}

\begin{document}
%
\conferenceinfo{ICMR}{2016 New York, USA}

\title{Scene-driven Retrieval in Edited Videos\\ using Aesthetic and Semantic Deep Features
}

%
%
%
%
%

\numberofauthors{1} 
%
\author{
\alignauthor
Lorenzo Baraldi, Costantino Grana and Rita Cucchiara\\
       \affaddr{Dipartimento di Ingegneria ``Enzo Ferrari'', Universit\`a degli Studi di Modena e Reggio Emilia}\\
       \affaddr{Via P. Vivarelli, 10, Modena MO 41125, Italy}\\
       \email{name.surname@unimore.it}
}

\maketitle
\begin{abstract}
This paper presents a novel retrieval pipeline for video collections, which aims to retrieve the most significant parts of an edited video for a given query, and represent them with thumbnails which are at the same time semantically meaningful and aesthetically remarkable. Videos are first segmented into coherent and story-telling scenes, then a retrieval algorithm based on deep learning is proposed to retrieve the most significant scenes for a textual query. A ranking strategy based on deep features is finally used to tackle the problem of visualizing the best thumbnail. Qualitative and quantitative experiments are conducted on a collection of edited videos to demonstrate the effectiveness of our approach.
\end{abstract}


\section{Introduction}
\begin{figure}[t]
    \centering
    \includegraphics[width=\columnwidth]{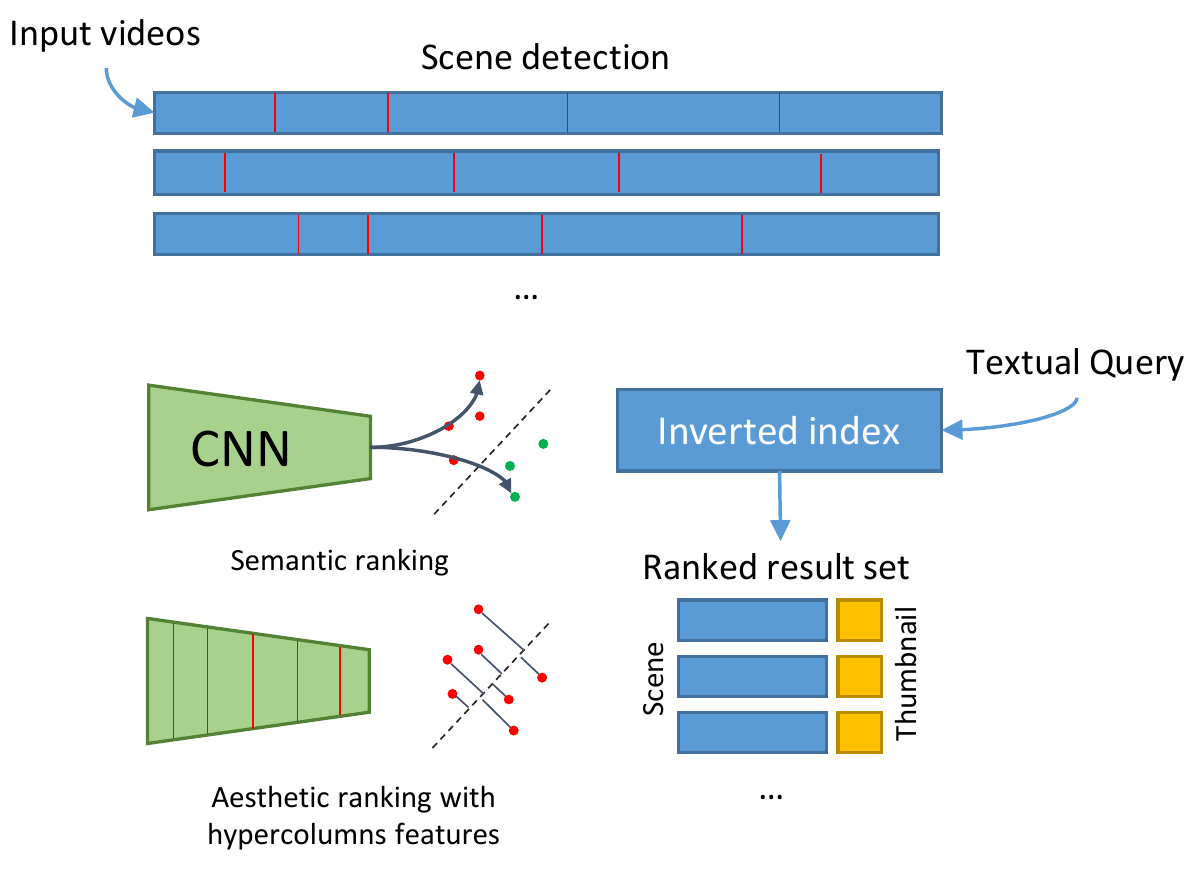}
    \caption{Overview of the proposed method. Given a collection of videos and a textual query, we retrieve a ranked list of the most significant parts (\textit{scenes}) according to both semantics and aesthetic quality. Each retrieved scene is presented with an appropriate thumbnail. }
    \label{fig:overview}
\end{figure}

Suppose to search for a given content in a large video collection, which contains long edited videos with different subjects and heterogeneous content, like a collection of documentaries or movies. In this context, users would like to have a quick overview of results, even with a low precision, but capable to give a glance of what can be associated with a query for a further manual refining. Examples are in \textit{advertisement} where video are re-used to find interesting sequences, in \textit{education} and \textit{edutainment} to enrich textual explanations with visual suggestions, in magazine \textit{editing}, in broadcast-to-web presentations, and also in web search engines.

Nowadays, retrieval is changing towards a greater focus on \textit{aesthetic quality}, a subjective aspect difficult to quantify. Datta \etal~\cite{datta2008image} assessed that modeling aesthetics of images is an important open problem, and it is still not solved. It concerns in general with the kind of emotions a picture arises in people, or more simply in beauty-related of images or videos.

This is an old story:  Plato, in \textit{Symposium}, discusses his idea of beauty, that comes from the perception of objects, their proportions, their harmony or unity among the parts, in the evenness of the line and purity of color. This Greek ideal permeates most of the occidental concepts of beauty and the current aesthetic theories, and affects as well theories on user interface designs and, recently, on retrieval too. Google, for instance, spent a large effort in changing the image search interface and the ranking, in order to convey not only the most relevant, abut also the most beautiful or fine results. Hongyi Li, associate product manager at Google, said that Google Images has been redesigned to provide ``a better search experience, faster, more beautiful and more reliable''\footnote{\url{https://googlewebmastercentral.blogspot.co.uk/2013/01/faster-image-search.html}}.
If image retrieval results are generally not only concerning the query but also ranked to have the more aesthetically valuable, this can be done also in video retrieval, where the complexity is higher. Moreover, also the granularity level could be changed: it is often the case, indeed, that long videos contain different parts and topics, hence an effective retrieval strategy should be able to recover the exact portion of the video the user is looking for.

In this paper we address the problem to provide both semantically and aesthetically valuable results of a query-by-text-retrieval in collections of long and heterogeneous video. Results are presented by thumbnails which recall the content of a video part associated with the specific search concept.
Our proposal addresses three common drawbacks of the existing video retrieval pipelines. First, we do not rely on manually provided annotations, like descriptions or tags, which are expensive and not always available, and exploit solely the visual and audio content of the video. Secondly, we let the user search inside a video with a finer granularity level. Lastly, once a set of candidate results has been collected, each should be presented to the user with a thumbnail which is coherent with the query and aesthetically pleasant. 
To lower the granularity level of searches, we temporally segment each video into a set of semantically meaningful sequences. This task, which is known in literature as \textit{scene detection}, results in a collection of scenes which have the property to be almost constant from a semantic point of view, and therefore constitute the ideal unit for video retrieval.

\section{Related work}
The process of producing thumbnails to represent video content has been widely studied. Most conventional methods for video thumbnail selection
have focused on learning visual representativeness purely from visual content~\cite{kang2005learn,rav2006making}; however, more recent researches have focused on choosing query-dependent thumbnails to supply specific thumbnails for different queries.
Craggs \etal~\cite{craggs2014thumbreels} introduced the concept that thumbnails are surrogates for videos, as they take the place of a video in search results. Therefore, they may not accurately represent the content of the video, and create an \textit{intention gap}, i.e. a discrepancy between the information sought by the user and the actual content of the video. To reduce the intention gap, they propose a new kind of animated preview, constructed of frames taken from a full video, and a crowdsourced tagging process which enables the matching between query terms and videos. Their system, while going in the right direction, suffers from the need of manual annotations, which are often expensive and difficult to obtain.

In~\cite{liu2011query}, instead, authors proposed a method to enforce the representativeness of a selected thumbnail given a user query, by using a  reinforcement algorithm to rank frames in each video and a relevance model to calculate the similarity between the video frames and the query keywords. Recently, Liu \etal~\cite{liu2015multi} trained a deep visual-semantic embedding to retrieve query-dependent video thumbnails. Their method employs a deeply-learned model to directly compute the similarity between a query and video thumbnails, by mapping them into a common latent semantic space.

On a different note, lot of work has also been proposed for video retrieval: with the explosive growth of online videos, this has become a hot topic in computer vision. In their seminal work, Sivic \etal proposed \textit{Video Google}~\cite{sivic2003video}, a system that retrieves videos from a database via bag-of-words matching. Lew \etal~\cite{lew2006content} reviewed earlier efforts in video retrieval, which mostly relied on feature-based relevance feedback or similar methods.

Recently, concept-based methods have emerged as a
popular approach to video retrieval. Snoek \etal~\cite{snoek2007adding} proposed a method based on a set of concept detectors, with the aim to bridge the semantic gap between visual features and high level concepts.  
In~\cite{ballan2015data}, authors proposed a video retrieval approach based on tag propagation: given an input video with user-defined tags, Flickr, Google Images and Bing are mined to collect images with similar tags: these are used to label each temporal segment of the video, so that the method increases the number of tags originally proposed by the users, and localizes them temporally. Our method, in contrast, does not need any kind of manual annotation, but is applicable to edited video only. 

\section{Visual-Semantic retrieval}
Given a set of videos $\mathcal{V}$, each decomposed into a set of scenes, and a query term $q$, we aim at building a function $r(q)$ which returns an ordered set of (video, scene, thumbnail) triplets. The retrieved scene must belong to the retrieved video, and should be as consistent as possible with the given query. Moreover, the returned thumbnail must belong to the given scene, and should be representative of the query as well as aesthetically remarkable.

To detect whether a (video, scene, thumbnail) triplet should correspond to a query, we exploit visually confirmed concepts found in the transcript, as well as a measure of aesthetic quality. We parse the transcript of a video to identify candidate concepts, like objects, animal or people. Then, for each concept a visual classifier is created \textit{on-the-fly} to confirm its presence inside the video, by means of an external corpus of images. Notice that when the transcript of video is not given, it can be easily replaced with the output of a standard speech-to-text software. \smallskip 

\begin{figure*}[tb]
    \centering
    \begin{subfigure}[m]{0.13\textwidth}
        \includegraphics[width=\textwidth]{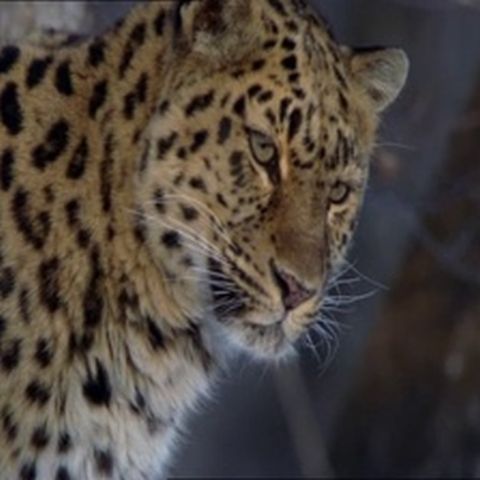}
        \caption{Input image}
    \end{subfigure}
    \begin{subfigure}{0.13\textwidth}
        \includegraphics[width=\textwidth]{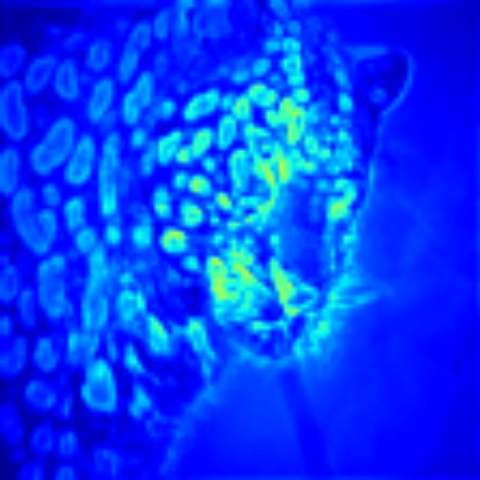}
        \caption{\texttt{conv1*}}
        \label{fig:conv1}
    \end{subfigure}
    \begin{subfigure}{0.13\textwidth}
        \includegraphics[width=\textwidth]{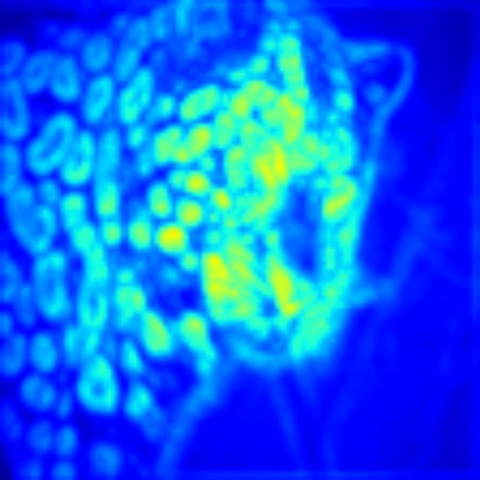}
        \caption{\texttt{conv2*}}
        \label{fig:conv2}
    \end{subfigure}
    \begin{subfigure}{0.13\textwidth}
        \includegraphics[width=\textwidth]{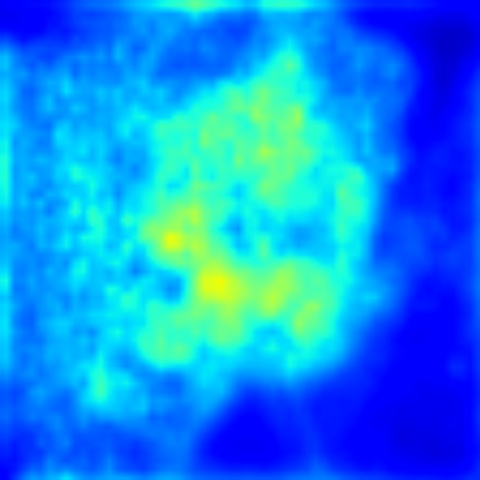}
        \caption{\texttt{conv3*}}
        \label{fig:conv3}
    \end{subfigure}
    \begin{subfigure}{0.13\textwidth}
        \includegraphics[width=\textwidth]{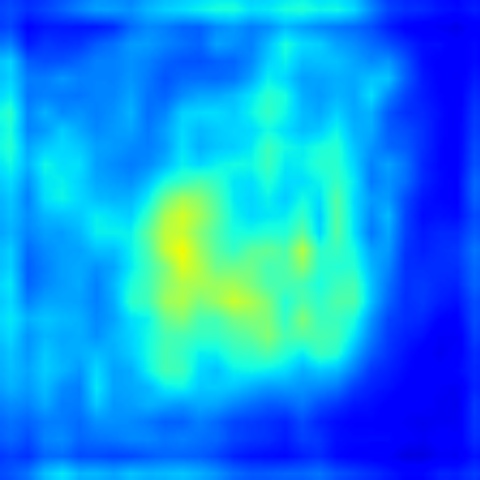}
        \caption{\texttt{conv4*}}
        \label{fig:conv4}
    \end{subfigure}
    \begin{subfigure}{0.13\textwidth}
        \includegraphics[width=\textwidth]{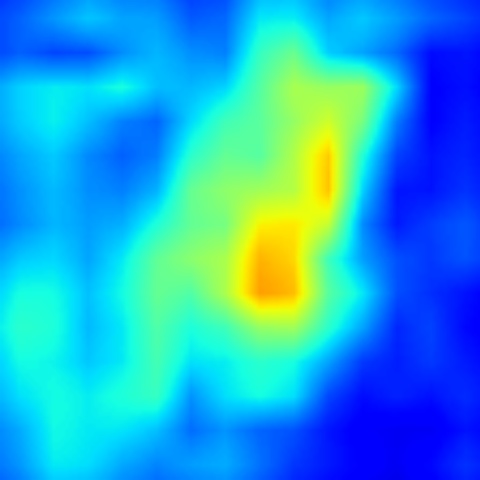}
        \caption{\texttt{conv5*}}
        \label{fig:conv5}
    \end{subfigure} \\
    
    \begin{subfigure}[m]{0.13\textwidth}
        \includegraphics[width=\textwidth]{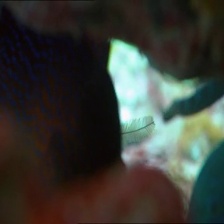}
        \caption{Input image}
    \end{subfigure}
    \begin{subfigure}{0.13\textwidth}
        \includegraphics[width=\textwidth]{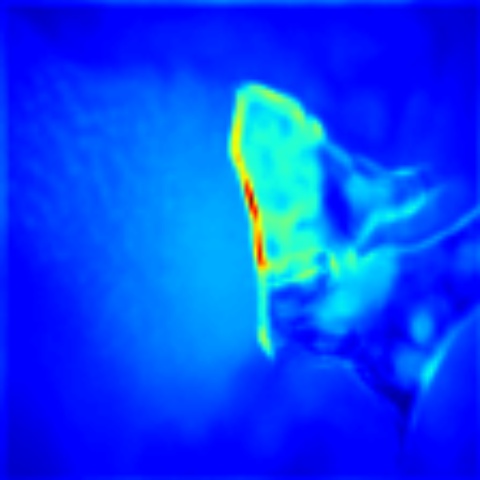}
        \caption{\texttt{conv1*}}
        \label{fig:conv11}
    \end{subfigure}
    \begin{subfigure}{0.13\textwidth}
        \includegraphics[width=\textwidth]{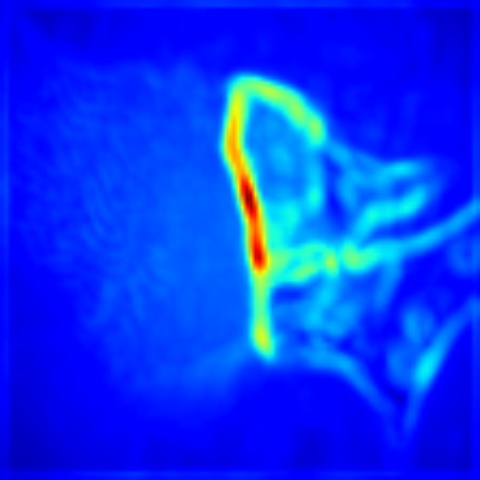}
        \caption{\texttt{conv2*}}
        \label{fig:conv22}
    \end{subfigure}
    \begin{subfigure}{0.13\textwidth}
        \includegraphics[width=\textwidth]{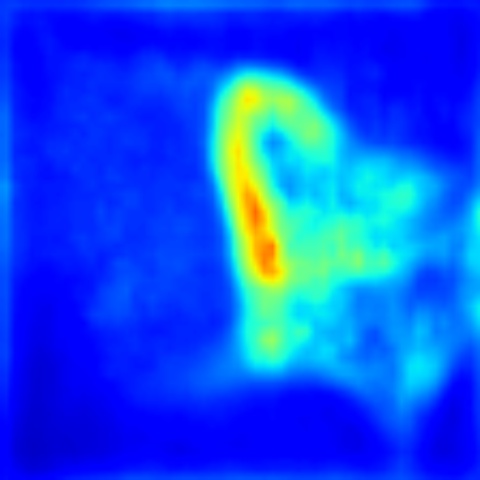}
        \caption{\texttt{conv3*}}
        \label{fig:conv33}
    \end{subfigure}
    \begin{subfigure}{0.13\textwidth}
        \includegraphics[width=\textwidth]{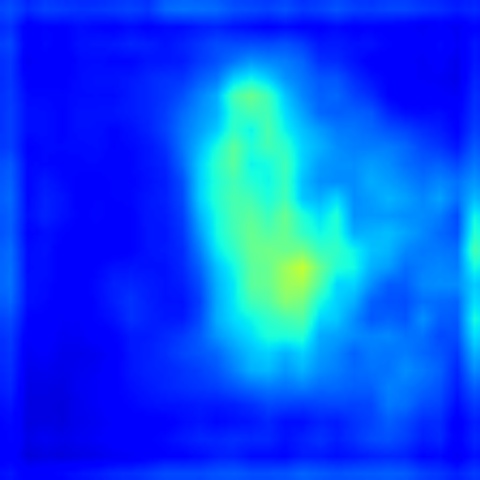}
        \caption{\texttt{conv4*}}
        \label{fig:conv44}
    \end{subfigure}
    \begin{subfigure}{0.13\textwidth}
        \includegraphics[width=\textwidth]{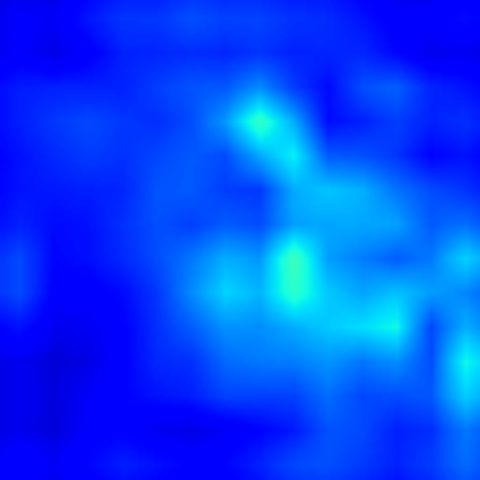}
        \caption{\texttt{conv5*}}
        \label{fig:conv55}
    \end{subfigure}
    \caption{Hypercolumn features extracted from two sample images. Each map represents the mean activation map over a set of layers: (b) and (h) are built using layers \texttt{conv1\_1} and \texttt{conv1\_2}, (c) and (i) with layers \texttt{conv2\_1} and \texttt{conv2\_2}; (d) and (j) with \texttt{conv3\_1}, \texttt{conv3\_2} and \texttt{conv3\_3}; (e) and (k) with \texttt{conv4\_1}, \texttt{conv4\_2}, and \texttt{conv4\_3}. Finally, (f) and (i) are built using layers \texttt{conv5\_1}, \texttt{conv5\_2} and \texttt{conv5\_3}. Best viewed in color.}
    \label{fig:hypercolumns}
\end{figure*}

\textbf{Scene detection}~To segment an input video into a set of coherent scenes, we apply the state-of-the-art algorithm described in~\cite{acm}. Given a ground-truth temporal segmentation of a set of videos,~\cite{acm} first runs a shot detector, then trains a Siamese Deep network to predict whether two shots should belong to the same temporal segment. Each branch of the Siamese network is composed by a Convolutional Neural Network (CNN) which follows the AlexNet architecture~\cite{krizhevsky12}, and whose penultimate layer is concatenated with features extracted from the transcript of the video. The overall network is trained using a contrastive loss function, which computes the distance between two input shots. In test phase, distances between shots provided by the Siamese network are arranged into a similarity matrix, wich is then used together with Spectral Clustering to obtain the final scene boundaries. \smallskip

\textbf{Semantic concept detection}~Sentences in the corpus are parsed and words annotated as \textit{noun}, \textit{proper noun} and \textit{foreign word} are collected with the Stanford CoreNLP part of speech tagger~\cite{de2006generating}. Each term is converted into its \textit{lemmatized} form, so that nouns in singular and plural form are grouped together. Due to the huge variety of concepts which can be found in the video collection, the video corpus itself may not be sufficient to train detectors for the visual concepts. Therefore, we mine images from the Imagenet database~\cite{deng2009imagenet}, which contains images from more than 40.000 categories from the WordNet~\cite{miller1995wordnet} hierarchy. Notice that our method, in principle, is applicable to any visual corpus, provided that it contains a sufficient large number of categories.

Each concept in WordNet is described by a set of words or word phrases (called \textit{synonim set}, or \textit{synset}). We match each unigram extracted from the text with the most semantic similar synset in a semantic space. In particular, we train a skip-gram model~\cite{mikolov2013efficient} on the dump of the English Wikipedia. The basic idea of skip-gram models is to fit the word embeddings such that the words in corpus can predict their context with high probability. Semantically similar words lie close to each other in the embedded space.

Word embedding algorithms assign each word to a vector in the semantic space, and the semantic similarity $S(u_1, u_2)$ of two concept terms $u_1$ and $u_2$ is defined as the cosine similarity between their word embeddings. For synsets, which do not have an explicit embedding, we take the average of the vectors from each word in the synset and $L$2-normalize the average vector. The resulting similarity is used to match each concept with the nearest Imagenet category: given a unigram $u$ found in text, the mapping function to the external corpus is as follows:
\begin{equation}
    M(u) = \argmax_{c \in \mathcal{C}} S(u,c)
\end{equation}
where $\mathcal{C}$ is the set of all concepts in the corpus. 

Having mapped each concept from the video collection to an external corpus, a classifier can be built to detect the presence of a visual concept in a shot. Since the number of terms mined from the text data is large, the classification step needs to be efficient, so instead of running the classifier on each frame of the video, we take the middle frame of each shot, using the shot detector in~\cite{apostolidis14}. At the same time, given the temporal coherency of a video, it is unlikely for a visual concept to appear in a shot which is far from the point in which the concept found in the transcript. For this reason, we run a classifier only on shots which are temporally near to its corresponding term, and apply a Gaussian weight to each term based on the temporal distance.

Images from the external corpus are represented using feature activations from pre-trained CNNs. In particular, we employ the 16-layers model from VGG~\cite{simonyan2014very}, pretrained on the ILSVRC-2012~\cite{ILSVRC15} dataset, and use the activations from layer \texttt{fc6}. Then, a linear probabilistic SVM is trained for each concept, using randomly sampled negative images from other classes; the probability output of each classifier is then used as an indicator of the presence of a concept in a shot.

Formally, given a shot $s$ which appears in the video at time $t_s$, and a unigram $u$ found in transcript at time $t_u$, the probability that $u$ is visually confirmed in $s$ is computed as:
\begin{equation}
P(s,u) = f_{M(u)}(s) e^{-\frac{(t_u - t_s)^2}{2\sigma_a^2}}
\end{equation} 
where $f_{M(t)}(s)$ is the probability given by the SVM classifier trained on concept $M(t)$ and tested on shot $s$. \smallskip

\begin{figure*}[tb]
    \centering
    \begin{subfigure}[m]{0.1\textwidth}
        \includegraphics[width=\textwidth]{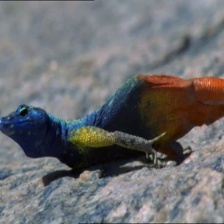}
    \end{subfigure}
    \begin{subfigure}{0.1\textwidth}
        \includegraphics[width=\textwidth]{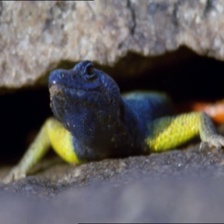}
    \end{subfigure}
    \begin{subfigure}{0.1\textwidth}
        \includegraphics[width=\textwidth]{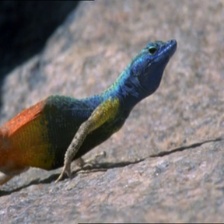}
    \end{subfigure}
    \begin{subfigure}{0.1\textwidth}
        \includegraphics[width=\textwidth]{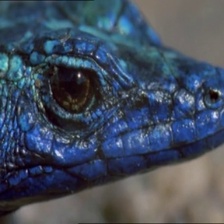}
    \end{subfigure}
    \begin{subfigure}{0.1\textwidth}
        \includegraphics[width=\textwidth]{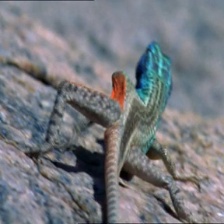}
    \end{subfigure}
    \begin{subfigure}{0.1\textwidth}
        \includegraphics[width=\textwidth]{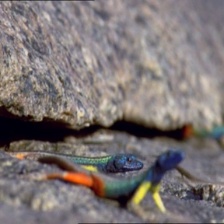}
    \end{subfigure}
    \begin{subfigure}{0.1\textwidth}
        \includegraphics[width=\textwidth]{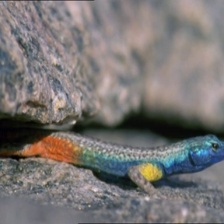}
    \end{subfigure}
    \begin{subfigure}{0.1\textwidth}
        \includegraphics[width=\textwidth]{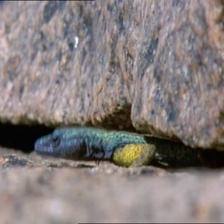}
    \end{subfigure}
    \begin{subfigure}{0.1\textwidth}
        \includegraphics[width=\textwidth]{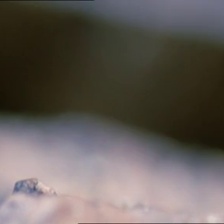}
    \end{subfigure}
    \caption{Ranking of a sample scene. Thumbnails with a centered and clearly visible animal are preferred against blurred and low-quality frames (best viewed in color).}
    \label{fig:ranking}
\end{figure*}

\begin{figure*}[bt]
    \centering
    \begin{subfigure}[m]{0.95\textwidth}
        \includegraphics[width=\textwidth]{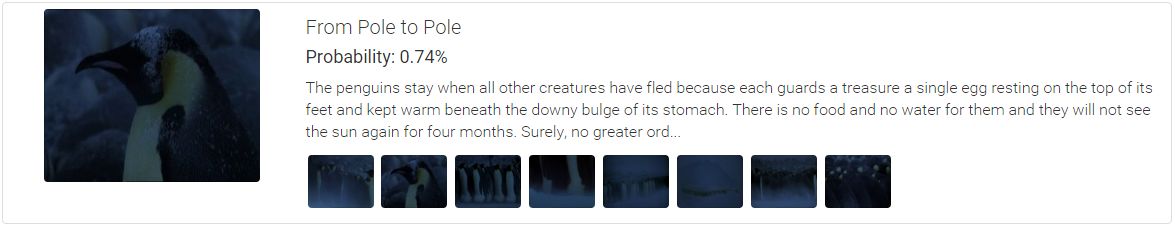}
        \caption{Result for query \textit{penguin}.}
    \end{subfigure} \\
    
    \begin{subfigure}[m]{0.95\textwidth}
        \includegraphics[width=\textwidth]{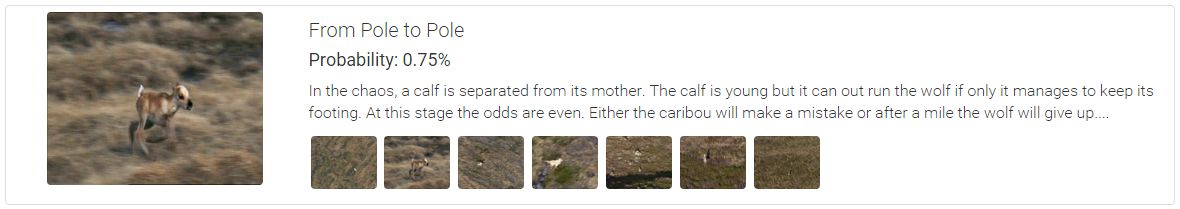}
        \caption{Result for query \textit{calf}.}
    \end{subfigure}
    \caption{Retrieval results. The same video is retrieved when searching for \textit{penguin} and for \textit{calf}, however, two different scenes are selected. Reported probability values correspond to $R_{scene}(q)$ in the paper.}
    \label{fig:query1}
\end{figure*}

\textbf{Aesthetic ranking}~The probability function defined above accounts for the presence of a particular visual concept in one shot, and is therefore useful to rank scenes given a user query. However, the thumbnail returned to the user should be visually representative as well. This requires a thumbnail selection step, which should account for low level characteristics, like color, edges and sharpness, as well as high level features, such as the presence of a clearly visible object in the center.

We claim that the need of low and high level features is an excellent match with the hierarchical nature of CNNs: convolutional filters, indeed, are known to capture low level as well as high level characteristics of the input image. This has been proved by visualization and inversion techniques, like ~\cite{zeiler2014visualizing} and~\cite{mahendran2015understanding}, which can visualize the role of each filter.

Being activations from convolutional filters discriminative for visual representativeness, a ranking strategy could be set up to learn their relative importance given a dataset of user preferences. However, medium sized CNNs, like the VGG-16 model~\cite{simonyan2014very}, contain more than 4000 convolutional filters, which produce as much activation maps. This makes the use of raw activations infeasible with small datasets: moreover, maps from different layers have different sizes, due to the presence of pooling layers. To get around with this, we resize each activation map to fixed size with bilinear interpolation, and average feature maps coming from the different layers, inspired by the Hypercolumn approach presented in~\cite{hariharan2015hypercolumns}. Moreover, since the the user usually focuses on the center of the thumbnail rather than its exterior, each maps is multiplied by a normalized gaussian density map, centered on the center of the image and with standard deviation $\sigma_b\cdot l$, where $l \times l$ is the size of the CNN input.

Following the VGG-16 architecture, we build five hypercolumn maps, each one summarizing convolutional layers before each pooling layer: the first one is computed with activation maps from layers \texttt{conv1\_1} and \texttt{conv1\_2}; the second one with \texttt{conv2\_1} and \texttt{conv2\_2}; the third with \texttt{conv3\_1}, \texttt{conv3\_2} and \texttt{conv3\_3}; the fourth with \texttt{conv4\_1}, \texttt{conv4\_2} and \texttt{conv4\_3}; the last with \texttt{conv5\_1}, \texttt{conv5\_2} and \texttt{conv5\_3}. An example of the resulting activation maps is presented in Fig.~\ref{fig:hypercolumns}: as it can be seen, both low level and high level layers are useful to distinguish between a significant and non significant thumbnail.

To learn the relative contribution of each hypercolumn map, we rank thumbnails from each scene according to their visual representativeness, and learn a linear ranking model. Given a dataset of scenes $\{ s_i \}_{i=0}^n$, each with a ranking $r_i^*$, expressed as a set of pairs $(d_i, d_j)$, where thumbnail $d_i$ is annotated as more relevant than thumbnail $d_j$, we minimize the following function:
\begin{equation}
\begin{aligned}
& \underset{\mathbf{w}, \mathbf{\epsilon}}{\text{minimize}}
& & \frac{1}{2} \|\mathbf{w}\|^2 + C\sum_{i,j,k}\epsilon_{i,j,k} \\
& \text{subject to}
& & \forall (d_i,d_j) \in r_1^* : \mathbf{w}\phi(d_i) \geq \mathbf{w}\phi(d_j) + 1 - \epsilon_{i,j,1} \\
& & & \ldots \\
& & & \forall (d_i,d_j) \in r_n^* : \quad  \mathbf{w}\phi(d_i) \geq \mathbf{w}\phi(d_j) + 1 - \epsilon_{i,j,n} \\
& & & \forall i, j, k : \epsilon_{i,j,k} \geq 0
\end{aligned}    
\label{eq:svm-rank}
\end{equation}

where $\phi(d_i)$ is the feature vector of thumbnail $d_i$, which is composed by the mean and standard deviation of each hypercolumn map extracted from the thumbnail itself. $C$ allows trading-off the margin size with respect to the training error. The objective stated in Eq.~\ref{eq:svm-rank} is convex and equivalent to that of a linear SVM on pairwise difference vectors $\phi(d_i)-\phi(d_j)$~\cite{joachims2002optimizing}. \medskip

\textbf{Retrieval}~Given a query $q$, we first match $q$ with the most similar detected concept $u$, using the Word2Vec embedding. If the query $q$ is composed by more than one words, the mean of the embedded vectors is used. Each scene inside the video collection is then assigned a score according to the following function:
\begin{equation}
R_{scene}(q) = \max_{s \in scene} \left(\alpha P(s,u) + (1-\alpha) \max_{d \in s} \mathbf{w}\phi(d)\right)
\label{eq:retrieval}
\end{equation}
where $s$ is a shot inside the given scene, and $d$ represent all keyframes extracted from a given shot. Parameter $\alpha$ tunes the relative importance of semantic representativeness and aesthetic beauty. The final retrieval results is a collection of scenes, ranked according to $R_{scene}(q)$, each one represented with the keyframe that maximizes the second term of the score.

From an implementation point of view, $P(s,u)$ can be computed offline for each unigram $u$ found in the transcript, for example with an inverted index. $\mathbf{w}\phi(d)$, as well, can be computed in advance for each key-frame, thus greatly reducing the computational needs in the on-line stage.

\section{Experimental results}
We evaluate the proposed method on a collection of 11 episodes from the \textit{BBC Planet Earth}\footnote{\url{http://www.bbc.co.uk/programmes/b006mywy}} series. Each video is approximately 50 minutes long, and the whole dataset contains around 4900 shots and 670 scenes. Each video is also provided with the transcript, and on the whole dataset a total of 3802 terms was extracted using the CoreNLP parser. Table~\ref{tab:dataset} reports some statistics on the dataset. Parameters $\sigma_a$ and $\sigma_b$ were set to 5 and 4.5 respectively, while $C$ was set to 3.

\subsection{Thumbnail selection evaluation}
Since aesthetic quality is subjective, three different users were asked to mark all keyframes either as aesthetically relevant or non relevant for the scene they belong to. For each shot, the middle frame was selected as keyframe. Annotators were instructed to consider the relevance of the visual content as well as the quality of the keyframe in terms of color, sharpness and blurriness. Each keyframe was then labeled with the number of times it was selected, and a set of $(d_i,d_j)$ training pairs was built according to the given ranking, to train our aesthetic ranking model.

For comparison, an end-to-end deep learning approach (\textit{Ranking CNN}) was also tested. In this case the last layer of a pre-trained VGG-16 network was replaced with just one neuron, and the network was trained to predict the score of each shot, with a Mean Square Error loss. Both the Ranking CNN model and the proposed Hypercolumn-based ranking were trained in a leave-one-out setup, using ten videos for training and one for test.
 
\begin{table}[tbp]
    \centering
    \begin{tabular}{|c|c|c|c|}
        \hline
        \textbf{Episode} & \textbf{Shots} & \textbf{Scenes} & \textbf{Unigrams} \\ \hline
        From Pole to Pole & 450 & 66 & 337 \\ \hline
        Mountains         & 395 & 53 & 339 \\ \hline
        Fresh Water       & 425 & 62 & 342 \\ \hline
        Caves             & 473 & 71 & 308 \\ \hline
        Deserts           & 461 & 65 & 392 \\ \hline
        Ice Worlds        & 529 & 65 & 343 \\ \hline
        Great Plains      & 534 & 63 & 336 \\ \hline
        Jungles           & 418 & 53 & 346 \\ \hline
        Shallow Seas      & 368 & 62 & 370 \\ \hline
        Seasonal Forests  & 393 & 57 & 356 \\ \hline
        Ocean Deep        & 470 & 55 & 333 \\ \hline
    \end{tabular}
    \caption{Statistics on the \textit{BBC Planet Earth} dataset.}
    \label{tab:dataset}
\end{table}

Table~\ref{tab:ranking} reports the average percent of swapped pairs: as it can be seen, our ranking strategy is able to overcome the Ranking CNN baseline and features a considerably reduced error percentage. This confirms that low and high level features can be successfully combined together, and that high features alone, such as the ones the Ranking CNN is able to extract from its final layers, are not sufficient. Figure~\ref{fig:ranking} shows the ranking results of a sample scene: as requested in the annotation, the SVM model preferred thumbnails with good quality and a clearly visible object in the middle.

\subsection{Retrieval results evaluation}
On a different note, we present some qualitative results of the retrieval pipeline. Figure~\ref{fig:query1} shows the first retrieved result when searching for \textit{penguin} and \textit{calf}, using $\alpha=0.5$. As it can be seen, our method retrieves two different scenes for the same video, based on the visually confirmed concepts extracted from the transcript. The same video, therefore, is presented with different scenes depending on the query. Moreover, selected thumbnails are actually representative of the query and aesthetically pleasant, when compared to the others available keyframes for those scenes. Depending on the query, it may also happen that the same scene is presented with two different thumbnails, as depicted in Fig.~\ref{fig:query2}: in this case the same scene was retrieved with query \textit{ant} and \textit{spider}; however, in the first case the selected thumbnail actually represents an ant, while in the second case a spider is selected, thus enhancing the user experience.


\begin{table}[tbp]
    \centering
    \begin{tabular}{|c|c|c|}
        \hline
        \multirow{2}{*}{\textbf{Episode}} & \multirow{2}{*}{\parbox{2.2cm}{\centering{\textbf{Ranking CNN}}}} & \multirow{2}{*}{\parbox{2.2cm}{\centering{\textbf{Hypercolumns Ranking}}}} \\
        & & \\ \hline
        From Pole to Pole & 8.23  & 4.10 \\ \hline
        Mountains         & 12.08 & 7.94 \\ \hline
        Fresh Water       & 12.36 & 8.11 \\ \hline
        Caves             & 9.98  & 8.76 \\ \hline
        Deserts           & 13.90 & 9.35 \\ \hline
        Ice Worlds        & 6.62  & 4.33 \\ \hline
        Great Plains      & 10.92 & 9.63 \\ \hline
        Jungles           & 12.28 & 7.43 \\ \hline
        Shallow Seas      & 10.91 & 6.22 \\ \hline
        Seasonal Forests  & 9.47  & 4.82 \\ \hline
        Ocean Deep        & 10.73 & 5.75 \\ \hline \hline
        \textbf{Average} & \textbf{10.68} & \textbf{6.95} \\ \hline
    \end{tabular}
    \caption{Aesthetic ranking: average percent of swapped pairs on the \textit{BBC Planet Earth} dataset (lower is better).}
    \label{tab:ranking}
\end{table}

\subsection{User evaluation}
\begin{figure*}[tb]
    \centering
    \begin{subfigure}[m]{0.85\textwidth}
        \includegraphics[width=\textwidth]{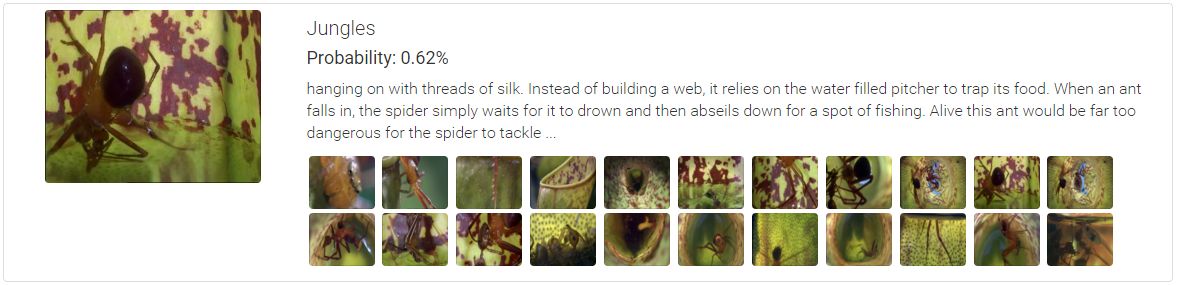}
        \caption{Result for query \textit{ant}.}
    \end{subfigure} \\
    
    \begin{subfigure}[m]{0.85\textwidth}
        \includegraphics[width=\textwidth]{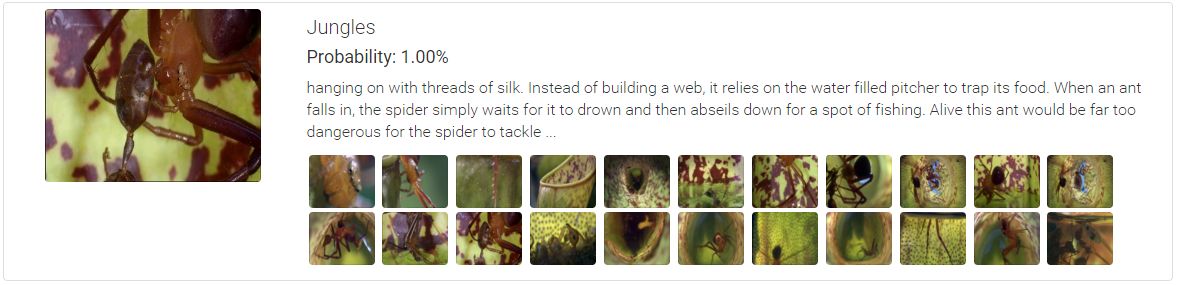}
        \caption{Result for query \textit{spider}.}
    \end{subfigure}
    \caption{Retrieval results. In this case the same scene from the same video is retrieved with two different queries (\textit{ant} and \textit{spider}), however, two different (and significant) thumbnails are selected. Reported probability values correspond to $R_{scene}(q)$ in the paper.}
    \label{fig:query2}
\end{figure*}

To quantitatively evaluate the ranking results and their effect on user experience, we conducted a user study with 12 undergraduate students. A demonstration and evaluation interface was built, in which the first three results returned by our method could be directly compared with three scenes retrieved with a full-text search inside the transcript, and presented with a random thumbnail different from the one selected by our system. As in Fig.~\ref{fig:query1} and \ref{fig:query2}, each retrieved scene was presented with the selected thumbnail, the corresponding transcription and with all the key-frames extracted from the scene. Users could also click on the thumbnail to watch the corresponding scene. 

Evaluators were asked to compare the provided result sets and vote the one they liked most, for a set of 20 queries. Results from our method were preferred to those provided by the baseline in the 82\% of cases, in the 15\% of evaluations they were said to be equivalent, while in the remaining 3\% of cases the baseline was preferred. 
The same queries were presented again replacing the thumbnails selected by our method with random ones. In this case the preferences were 12\% for the baseline and 57\% for our proposal, while in the 31\% of cases results were evaluated as equivalent. 

This confirms the importance of selecting appropriate thumbnails when dealing with casual users (the students didn't have any real goal, nor were particularly interested in the queries we provided). One of the conclusions we can draw from this tests is that the presentation of the results may strongly influence the feeling of "correctness" of the retrieved results.



\section{Conclusions}
We presented a novel video retrieval pipeline, in which videos are decomposed into short parts (namely scenes), that are used as the basic unit for retrieval. A score function was proposed to rank scenes according to a given textual query, taking into account the visual content of a thumbnail as well as its aesthetic quality, so that each result is presented with an appropriate keyframe. Both the semantics and the aesthetics were assessed using features extracted from Convolutional Neural Networks, and by building on-the-fly classifiers for unseen concepts. Our work has been evaluated both in qualitative and quantitative terms, and results in enhanced retrieval results and user experience.
%
\bibliographystyle{abbrv}
\balance
\bibliography{sigproc}  

\end{document}